\documentclass[11pt,a4paper]{article}
\usepackage[hyperref]{acl2021}
\usepackage{times}
\usepackage{latexsym}

\usepackage{microtype}

\aclfinalcopy 

\title{Detecting Unseen Multiword Expressions in American Sign Language}

\author{Lee Kezar \\
  University of Southern California \\
  \texttt{lkezar@usc.edu} \\\And
  Aryan Shukla \\
  University of Southern California \\
  \texttt{aryanshu@usc.edu} \\}

\date{}

\begin{document}
\maketitle
\begin{abstract}
Multiword expressions present unique challenges in many translation tasks. In an attempt to ultimately apply a multiword expression detection system to the translation of American Sign Language, we built and tested two systems that apply word embeddings from GloVe to determine whether or not the word embeddings of lexemes can be used to predict whether or not those lexemes compose a multiword expression. It became apparent that word embeddings carry data that can detect non-compositionality with decent accuracy. 

\end{abstract}

\section{Introduction}

Translating signed languages, such as American Sign Language (ASL), has been challenging for machine translation systems. This is particularly problematic considering that there are between several hundred thousand and several million people in the United States who are deaf, hard of hearing or might otherwise rely on ASL as their most convenient method of communication. That said, ASL translation requires many more steps in translation \citep{bragg2019sign}. One fundamental processing task involves the detection of multiword expressions (MWEs) in ASL. 

Translating MWEs is a unique challenge (compared to translating single words) because their meaning is frequently derived from an idiomatic use of multiple lexemes. At the same time, those lexemes can be used individually and non-idiomatically, which is a much more straightforward task for translation. This ambiguity motivates the detection of MWEs, so as to avoid deep misunderstandings in language processing. The present research leverages these observations to detect compound MWEs based on their non-compositionality. While other MWEs such as multiword entities (e.g. \textit{long short-term memory network}) can also exhibit non-compositionality, \citet{constant-etal-2017-survey} present three unique challenges for compounds:
\begin{enumerate}
    \item [1.] \textbf{Inconsistency} There is a wide range of diversity in the structure in which compounds can appear.
    \item [2.] \textbf{Contextual Dependence} Frequently, context indicates that the target set of words are an MWE, as opposed to linguistic identifiers that surround compounds.
    \item [3.] \textbf{Non-contiguity} In languages with non-obvious separations between lexemes, such as logographic and signed languages (e.g. written Chinese and ASL, respectively), compounds may be mistakenly parsed as separate tokens.
\end{enumerate}

For ASL in particular, there is the additional challenge of being a very low resource environment, complicating the use of neural methods like transformers \citep{bragg2019sign}. In the absence of a labeled dataset of ASL MWEs, we present a method for detecting separated English compounds (e.g. ``home work") that leverages word embeddings and definitions, both of which can be adapted to low-resource environments \citep{DBLP:journals/corr/abs-1809-03633}. To evaluate this method, we form two hypotheses:

\begin{enumerate}
    \item[\textbf{H1}] Co-occurrence across related contexts will be an effective signal for detecting non-contiguous compounds.
    \item[\textbf{H2}] The definition of each lexeme that composes a compound MWE may contain additional information about the context in which a compound might appear.
\end{enumerate}

We find that our methods can detect compounds with high recall, indicating that many compounds are non-compositional in nature, and that definition-based detection is effective for low-resource environments. Future work will study the generalizability of these methods for glossed ASL.

\section{Related Work}
Non-compositionality is one of the several properties of MWEs that can be used for detection. Non-compositionality describes how the meaning of a MWE may not be correlated with the individual components that make it up. This property ``is generally leveraged by models based on vector-space semantic similarity” \citep{constant-etal-2017-survey}. Many compounds can be non-compositional, and thus, this property can be applied to detect them. In terms of word embeddings, if the lexemes of a compound co-occur regardless of their MWE status, then they should have relatively similar word embeddings \citep{kiela-clark-2014-systematic}. For example, ``video lag" can be expected to have high similarity, whereas a non-compositional MWE, such as ``jet lag" should not have high similarity. Embeddings also have the advantage of generalizing to low-resource environments, such as ASL. Note that non-compositionality has the limitation of non-universality; that is, not all compounds are necessarily non-compositional \citep{constant-etal-2017-survey}.

There have been several, often successful, attempts at accomplishing MWE detection through the application of vector space models. Two methods are the most notable: One method, presented by \citet{kiela-clark-2014-systematic} to determine the compositionality of a phrase by substituting synonyms in a phrase forms a conceptual foundation for the potential for word embeddings to suggest compositionality. The authors compare MWEs by computing the distance between a MWE’s vector and the vectors of version of the same MWE with synonyms substituted for individual lexemes in that MWE. Their results support the intuition that word embeddings can predict whether substituted versions of MWE are ``meaningful” or not. Their results suggest that word embeddings may carry information about the compositionality of a phrase. Another method, presented by \citet{salehi-etal-2015-word} forms a conceptual foundation for comparing individual word embeddings. This work operates on principles similar to ours, except it leverages part-to-whole relationships, such as \textit{snow vs. snowball}, instead of part-to-part relationships, such as \textit{snow vs. ball}. One limitation of this work is that it does not study detection directly, so it is unclear how it would perform with non-MWEs. ASL translation is one occasion where this presents a problem; compounds will often be expressions like ``RED CUT" (``tomato'') with no specific signal to suggest that the two words should be understood together as a compound. Our methods, however, build on the work of \citet{salehi-etal-2015-word} to determine the potential of a system that is blind to the frame of a potential compound and is capable of simply running on all groups of cooccurring lexemes to determine whether or not the set of lexemes might be a compound.

\begin{table*}[t]
    \centering
    \begin{tabular}{|c | c c c|}
    \hline
     Method & Recall & Precision & F1 Score \\
     \hline
     word similarity & 0.840 & 0.596 & 0.697 \\ 
     \hline
     definition similarity & 0.847 & 0.481 & 0.613 \\
     \hline
     definition content similarity & 0.859 & 0.483 & 0.619 \\
     \hline
    \end{tabular}
    \caption{Accuracy of methods using random word pair negative samples.}
    \label{tab:rand}
\end{table*}
\begin{table*}[t]
    \centering
    \begin{tabular}{|c | c c c|}
    \hline
    Method & Recall & Precision & F1 Score \\
    \hline
    word similarity & 0.840 & 0.754 & 0.795 \\ 
    \hline
    definition similarity & 0.847 & 0.527 & 0.649 \\
    \hline
    definition content similarity & 0.859 & 0.532 & 0.657 \\
    \hline
    \end{tabular}
    \caption{Accuracy of methods using co-occurring word pair negative samples.}
    \label{tab:freq}
\end{table*}

\section{Method}

To test our hypotheses, we introduce three different scores based on the cosine similarity between different elements of the two lexemes of a potential compound. Our first method states that if a word is a compound, then the embeddings of the lexemes that compose it will have high cosine similarity (``word similarity"). The second method leverages the capacity of definitions of each lexeme to contain contextualizing terms. Thus, we predicted that the embedding of the definitions will also have high cosine similarity (``definition similarity"). We define a definition embedding as the elementwise sum of the embeddings of words in a provided definition. Finally, we optionally remove stop words from definitions to improve consistency (``definition content similarity"). 
 
To test each method, we used data from the Large Database of English Compounds (LADEC) to provide a sample of 8956 compounds, mixed with an equal number of random word pairs and an equal number of frequently co-occurring word pairs from the Brown Corpus \citep{Gagne2019}. The scores, then, must distinguish between ``home work" (compound), ``home play'' (random), and ``home chef'' (frequent bigram). We compare the distribution of random word pairs to known compounds to determine an appropriate threshold for classification, a step that generalizes to low-resource settings where a modest list of compounds can be procured by hand. After establishing this maximum-effectiveness threshold, we label each unseen pair and compute overall performance.

\section{Results}

We evaluate our methods by computing recall, precision, and F1 scores for each score's ability to distinguish LADEC compounds from negative samples. Based on the aforementioned threshold calibration, it was decided that the threshold for compound is 0.78 for word similarity, 0.90 for definition similarity, and 0.46 for definition content similarity. Any value returned below the threshold for each method could be considered as a “compound” judgement whereas any value returned above the threshold could be considered a “not a compound” judgement. It is important to note that these methods are making compound judgements based solely on the non-compositionality of compounds, so it is not expected that all compounds, especially compositional compounds, will be detected. The findings are summarized in Tables \ref{tab:rand} and \ref{tab:freq}. 

\section{Discussion}

We find substantial support for H1, and less for H2. Because the similarity of two compound-forming lexemes in a vector space proved to be decently accurate in analyzing whether a word is a compound, H1 demonstrated that vector space embeddings that are made through the distributional hypothesis, with words that are likely to appear together being given embeddings that are closer together, are a useful way to predict compounds. H2 was supported less, since the accuracy of the second method was not higher than the accuracy of the first method, so the splitting of a word into the individual words that composed its definition did not provide important context about the appearance of the word in a sentence or paragraph. In fact, since Method 1 proved to be more accurate while simply plotting the locations of each lexeme rather than the sum of the words in each lexeme’s first definition, our experiments suggest that the 100-dimensional embeddings actually provide more significant information about the context of a lexeme that allows us to better judge a compound’s compositionality than the words in a lexeme’s definition. We also discovered that removing stop words hardly made the system more accurate in correctly determining whether or not a word was a compound, which may suggest that the stop words as a part of the word’s definition are only marginally important. These generalized results stayed consistent whether we used the random word pairs or the cooccurring word pairs as the negative samples for comparison. However, the higher accuracy for co-occurring word pair negative samples suggests that compounds may be easier to detect in contextualized language because meaningless word pairs with no context are more likely to be non-compositional like some compounds themselves. The recall remained the same for both the random word pair and cooccuring word pair negative samples, since recall compares true positives to all positives and the positive samples in our tests were the same alongside both sets of negative samples (the compounds presented in \citet{Gagne2019}). The lower precision on the definition-based methods, however, suggest that they cast a wider net in detecting non-compositionality because the individual words of the definitions summed do not contain the same key contextualizing data as the embeddings of words themselves. 

Our results show strong evidence that word embeddings can be applied to detect compositionality in lexemes for detecting compounds. However, our results are certainly limited. The word embeddings that we used are from \citet{pennington2014glove}. This system groups word embeddings by definition and synonyms, substructures based on basic knowledge graphs, and more, but may not be directly intended to detect compositionality between two lexemes, since that requires context. Additionally, our models have only been adjusted on and applied to a very limited set of data; we are dependent on the Brown corpus for random samples and the LADEC for true positive compounds, and each of those datasets contain flaws based on the likelihood for certain word pairs to appear based on the news context of Brown’s corpus \citep{bird_natural_2009}. Furthermore, our results should be applied and understood with one key distinction; these systems were detected on a set of all compounds in LADEC, which includes both compositional and non-compositional compounds. Because the system is premised on detecting non-compositionality and not necessarily compositional compounds, this distinction may greatly affect our dataset. Because the LADEC contains data about the likely compositionality of various compounds, it is possible to attempt to test these systems on only non-compositional compounds, but the system will ultimately be applied to detect all compounds in our ultimate application in ASL translation, so we tested it on all compounds. This is significant to take note of when analyzing our data, but our conclusions remain valid due to the simple fact that the population of compounds in LADEC will, in general, be more non-compositional than average word pairs since many compounds inherently exist because of their non-compositionality. 

\section{Future Work and Conclusion}

The distributional hypothesis used in producing word embeddings provides a useful way to detect compounds through closeness of embedding locations, which may provide information on how descriptive one lexeme is of the other, and thus, determine compositionality. Individual lexeme definitions do not contain important context information that embeddings do not already take into account that would make them any more useful in determining the compositionality of a pair of lexemes, and stop words seem to have no significant effect in changing that usefulness. Ultimately, it has become clear that word embeddings have proven to be a useful tool in determining the compositionality of a pair of lexemes. 

We hope that these findings prove to be useful in determining a more robust MWE detection mechanism that is able to find multiword expressions out of a set of sentences. While this work has been done assuming that inputted MWEs have already been split into their compositional lexemes, this tool can be applied more practically if it can work alongside a method to isolate lexemes in a MWE, even if they are combined in the form of one word. Ultimately, we hope to apply this mechanism to ASL translation, where compounds are already separated into individual morphemes. Additionally, attempting to test this system on compounds that are either explicitly judged to be non-compositional can give us better insight into the logic and reasons for the success of these methods. 


\bibliography{custom}
\bibliographystyle{acl_natbib}

\end{document}